\documentclass[10pt, a4paper]{article}

\usepackage{amsmath}
\usepackage{xcolor} 
\usepackage{multirow}
\usepackage{graphicx}
\usepackage{booktabs} 
\usepackage{siunitx}
\newcommand{\rotlabel}[1]{\rotatebox[origin=c]{90}{\scriptsize #1}}
\newcommand{\aspectsep}{\cmidrule(lr){3-7}}

\usepackage[final]{lrec2026} 

\title{Rubric-Guided Fine-tuning of SpeechLLMs for Multi-Aspect, Multi-Rater L2 Reading-Speech Assessment}

\name{Aditya Kamlesh Parikh, Cristian Tejedor-Garcia, Catia Cucchiarini, Helmer Strik} 

\address{Centre for Language Studies, Radboud University, The Netherlands \\
        \{aditya.parikh, cristian.tejedorgarcia, catia.cucchiarini, helmer.strik\}@ru.nl\\}

\abstract{
Reliable and interpretable automated assessment of second-language (L2) speech remains a central challenge, as large speech-language models (SpeechLLMs) often struggle to align with the nuanced variability of human raters. To address this, we introduce a rubric-guided reasoning framework that explicitly encodes multi-aspect human assessment criteria: accuracy, fluency, and prosody, while calibrating model uncertainty to capture natural rating variability. We fine-tune the Qwen2-Audio-7B-Instruct model using multi-rater human judgments and develop an uncertainty-calibrated regression approach supported by conformal calibration for interpretable confidence intervals.
Our Gaussian uncertainty modeling and conformal calibration approach achieves the strongest alignment with human ratings, outperforming regression and classification baselines. The model reliably assesses fluency and prosody while highlighting the inherent difficulty of assessing accuracy. Together, these results demonstrate that rubric-guided, uncertainty-calibrated reasoning offers a principled path toward trustworthy and explainable SpeechLLM-based speech assessment.
 \\ \newline \Keywords{SpeechLLM, L2 Reading Speech, Multi-Aspect Assessment, SpeechLLM Fine-tuning, Uncertainty Modeling} }

\begin{document}

\maketitleabstract

\section{Introduction}

Reading is a foundational skill for learning, communication, and participation in society. It includes multiple aspects: readers must pronounce words accurately \citep{newell2020oral}, maintain a natural temporal flow to ensure fluency \citep{kuhn2003fluency}, and convey appropriate phrasing and emphasis to reflect prosody \citep{schwanenflugel2004becoming}. These dimensions interact in complex ways. For instance, prosodic phrasing can mask or amplify word-level inaccuracies \citep{zechner2015automated}.  

Many readers, including children developing basic reading skills \citep{schwanenflugel2004becoming}, L2 learners acquiring phonological and rhythmic competence \citep{sleeman2022identification}, and adults encountering unfamiliar vocabulary or orthographies \citep{chang2020relationships}, struggle to achieve fluent, accurate, and prosodically natural reading. Early, high-quality feedback is essential for enabling teachers and clinicians to identify learners' needs and adjust instruction \citep{gronli2024teachers}.

Nevertheless, expert assessment and feedback are costly, time-consuming, often inconsistent across raters \citep{smith2019study}, and suffer from variability in severity and scale usage (leniency, harshness, central tendency, and halo effects) and from drift over time due to fatigue or anchoring \citep{neittaanmaki2024communal}. Ensuring inter- and intra-rater reliability across raters and sessions remains challenging, especially for child and L2 speech, where accent variation and atypical prosody further reduce agreement \citep{ishikawa2023effects}. In particular, raters tend to agree more on pronunciation accuracy than on fluency or prosody \citep{van2024they}.

Diverse modeling approaches have been explored to develop automatic systems for speaking and reading assessment. Early work in the 1990s relied on statistical models such as Hidden Markov Models (HMMs) and Gaussian Mixture Models (GMMs), which primarily assessed pronunciation in reading speech using posterior probabilities or Goodness of Pronunciation (GOP) scores \citep{kim1997automatic,witt2000phone,witt2000use}. Subsequent studies broadened the scope to fluency and prosody, introducing timing- and pitch-based features \citep{wang2024improving}.

With the advent of deep neural networks, pronunciation assessment research began emphasizing calibration to ensure that a model’s predicted confidence aligns with the reliability of its assessments \citep{evanini2018improvements}. Later, text-based Language Models (LMs) such as BERT \citep{devlin2019bert} and RoBERTa \citep{liu2019robertarobustlyoptimizedbert} were applied to automated scoring using Automatic Speech Recognition (ASR) transcripts; however, their dependence on textual transcriptions and lack of acoustic awareness limited their ability to capture fluency and prosody effectively.
The emergence of pretrained acoustic models such as wav2vec 2.0 \citep{baevski2020wav2vec} addressed this gap, improving performance on tasks like mispronunciation detection and diagnosis \citep{cao24b_interspeech,parikh25_interspeech,parikh25b_interspeech,phan25_interspeech}. In parallel, the rise of multimodal Large Language Models (LLMs) has extended natural language processing beyond text and inspired similar progress in the speech domain. For instance, models such as SpeechLLaMA \citep{wu2023decoder}, SALMONN \citep{tang2023salmonn}, VOILA \citep{wang2023viola}, and Qwen-Audio \citep{chu2023qwen} expanded text-based architectures by integrating acoustic and linguistic representations. These SpeechLLMs demonstrated strong performance on general audio understanding tasks such as speech recognition, translation, audio captioning, and spoken question answering, yet they still lacked the capability to follow detailed natural-language instructions.

A newer generation of instruction-tuned SpeechLLMs has recently emerged, including Qwen2-Audio-Instruct \citep{chu2024qwen2}, GAMA \citep{ghosh-etal-2024-gama}, and Audio Flamingo 2 \citep{pmlr-v267-ghosh25b}. These models combine large-scale audio–text pretraining with instruction tuning, allowing them to reason over spoken input and generate structured responses directly from raw speech. While such instruction-tuned models hold significant potential for rubric-guided and explainable assessment of L2 speech proficiency, they remain underexplored in the literature. Recent work has begun to highlight both their promise and limitations. For instance, \citet{parikh2025zero} and \citet{ma25b_interspeech} investigated the Qwen2-Audio-Instruct model under distinct settings. \citet{parikh2025zero} observed that rubric-based SpeechLLMs tend to produce overly generous scores in zero-shot conditions, reflecting a “niceness bias” inherited from instruction tuning that discourages low ratings even for poor-quality speech. Their study also introduced a multi-aspect assessment framework, allowing simultaneous scoring of complementary proficiency dimensions such as accuracy, fluency, prosody, and sentence completeness. In contrast, \citet{ma25b_interspeech} demonstrated that supervised fine-tuning can effectively mitigate this bias, yielding more consistent and reliable proficiency predictions. Nonetheless, their framework was limited to single-score regression and classification setups and did not account for inter-rater variability or predictive uncertainty, two critical factors for ensuring fairness and reliability in automated assessment.

Building upon these findings, we adopt the Qwen2-Audio-Instruct model as the foundation of our study, utilizing its instruction-following capacity with paired audio inputs and descriptive textual rubrics for assessment.
The model is fine-tuned in a multi-aspect manner to infer holistic scores along three complementary dimensions: accuracy (degree of mispronunciation), fluency (smoothness and coherence of delivery), and prosody (intonation, rhythm, and stress). We will share our code repository upon acceptance.

To systematically examine whether incorporating additional configurations can improve robustness and alignment with human judgments, we design five state-of-the-art (SOTA) modeling strategies of increasing complexity based on the related literature. (1) Discrete Classification (\textbf{DiCl}) treats proficiency scoring as a categorical prediction task \citep{xi2012comparison}. However, this uniform treatment of errors disregards the ordinal relationship between categories, which is crucial in assessment tasks where the severity of misclassification depends on the distance between true and predicted proficiency levels; (2) Single-Rubric Regression (\textbf{SRR.M}) formulates the scoring as continuous prediction using mean squared error (MSE) \citep{chen2018automated}, aligning better with human rating scales and capturing finer performance differences; (3) Multi-Rubric Regression (\textbf{MRR.M}) jointly predicts multiple rubrics simultaneously with MSE \citep{do23b_interspeech}, enabling shared representation learning in multiple aspects such as accuracy, fluency, and prosody; (4) Multi-Rubric Regression with Gaussian Negative Log-Likelihood (GNLL), with the acronym \textbf{MRR.G}, replaces MSE with GNLL to model prediction uncertainty \citep{kendall2017uncertainties}; and (5) Multi-Rubric Multi-Rater Regression with GNLL and Conformal Prediction (\textbf{MRR.GC}) incorporates multiple human ratings and applies Conformal Prediction \citep{angelopoulos2021gentle,braun2025multivariate} to generate calibrated confidence intervals.
Among these, the last two configurations (MRR.G and MRR.GC) represent novel contributions to the field of automated L2 speech assessment. To the best of our knowledge, this is the first study to integrate Gaussian uncertainty modeling and conformal calibration within a multi-rater supervision framework for multi-aspect assessment using a rubric-guided fine-tuned SpeechLLM. This design advances the SOTA by jointly addressing score reliability, fairness, and explainability, three critical yet previously underexplored dimensions in SpeechLLM-based proficiency assessment.

This leads us to our research question (RQ):
\textit{To what extent can a SpeechLLM approximate human ratings in multi-aspect (accuracy, fluency, and prosody) performance assessment of L2 reading speech?}

\section{Methodology}

In this section, we describe the experimental setup for our sentence-level speech assessment framework, which leverages a multimodal (speech and text) SpeechLLM fine-tuned to predict rubric-based scores for multi-aspect (accuracy, fluency, and prosody) assessment.
The task was framed either as classification (five discrete levels: Very Poor to Excellent) or as regression (continuous scores on a 1–10 scale). While classification provides interpretable, rubric-aligned decisions, regression enables finer granularity and captures subtle variations in human ratings \citep{xi2012comparison}.
We fine-tuned the Qwen2-Audio-7B-Instruct\footnote{\url{https://huggingface.co/Qwen/Qwen2-Audio-7B-Instruct}} model using Low-Rank Adaptation (LoRA) \citep{hu2022lora} for parameter-efficient adaptation while preserving the model’s pre-trained capabilities.
The following subsections describe the model architecture, dataset, training setup, optimization loss functions, and the assessment protocol.

\subsection{Model Architecture}

We built upon the Qwen2-Audio-7B-Instruct model, a 7B-parameter multimodal transformer-based SpeechLLM pre-trained on large-scale audio–text pairs for conditional generation tasks. The model integrates an audio encoder with a text decoder, enabling it to process interleaved audio and textual instructions. For speech assessment, a lightweight scoring head was added to project the hidden representations from the final transformer layer to output predictions. We explored two variants: (i) a classification head trained with cross-entropy loss for discrete 5-level scoring, and (ii) a regression head trained with MSE or GNLL loss to predict continuous scores. 

To enable parameter-efficient fine-tuning, LoRA was applied to all linear layers of the base model. The key hyperparameters were a rank of \(r = 32\), a scaling factor of \(\alpha = 32\), and a dropout rate of \(0.1\), with rank-stabilized LoRA (rsLoRA) enabled. 
This configuration resulted in approximately 10 million trainable parameters (\(\approx1.2\%\) of the total), focusing the optimization on low-rank update matrices while keeping the original model weights frozen.

\subsection{Dataset}

We used the publicly available dataset SpeechOcean762 \citep{zhang21x_interspeech}, a widely adopted benchmark for automatic pronunciation and speaking assessment for research. It contains 5000 English read-speech utterances, divided into 2500 for training and 2500 for testing. We fine-tuned our model on the training split and assessed the test split.
The corpus includes recordings from both child and adult speakers whose native language is Mandarin Chinese (L1), reading English (L2). Each utterance was independently evaluated by five expert raters along sentence, word, and phoneme level aspects. Scores were assigned on a 1–10 scale following the official annotation protocol defined in the original dataset publication. This dataset is particularly suited for our study as it provides multi-rater, multi-aspect annotations, enabling analysis of both inter-rater variability and multi-dimensional scoring behavior. In this work, we focus on sentence-level scoring (accuracy/fluency/prosody) to study rubric-guided utterance-level assessment under multi-rater supervision. We exclude Completeness due to an unclear rubric definition for our setup. Scores are skewed toward mid- to high ratings, potentially biasing predictions toward the central range.

In a multi-aspect assessment of speech, accuracy is measured as the pronunciation correctness of the spoken sentence. A score of 10 corresponds to excellent pronunciation with no noticeable mispronunciations (near-native articulation), whereas 1 indicates completely unintelligible speech or absence of voice. 
Fluency evaluates the temporal smoothness and coherence of speech, focusing on pauses, repetitions, and stammering. A score of 10 reflects coherent, uninterrupted delivery with natural pacing, while 1 denotes inability to read the sentence as a whole or no voice.
Finally, Prosody measures the intonation, rhythm, and speaking rate, capturing the naturalness and expressiveness of speech. A score of 10 represents correct intonation with stable rhythm and speed, sounding natural and engaging, and 1 indicates speech too stammered to evaluate or the absence of voice. 

The scoring rubrics for all three aspects follow the definitions provided in the SpeechOcean762 paper (\citep{zhang21x_interspeech}, page 3).

\subsection{Training Procedure}

Fine-tuning of the Qwen model was conducted using the Hugging Face Transformers library with a custom data collator and trainer \citep{wolf-etal-2020-transformers} as it provides flexibility in handling paired audio–text inputs and rubric-based labels. 
Each utterance was resampled to 16 kHz, converted to mono audio, and paired with a textual task prompt containing only rubric instructions (no target transcript), so the model assesses directly from audio rather than transcript comparison.

The input was formatted as a conversational prompt following the model’s chat template, consisting of a user message that included the audio segment and a rubric-based instruction (e.g., “Score sentence-level accuracy on a scale from 1 to 10,” accompanied by detailed rubric descriptions). Since the objectives included both regression and classification, no generation prompt was added, as the model was trained to produce direct scalar or categorical predictions rather than generated text. 
For optimization, the AdamW optimizer was used with a learning rate of \(2 \times 10^{-5}\), a weight decay of \(0.01\), and a constant learning rate schedule (no warm-up). The per-device batch size was set to \(1\), with gradient accumulation steps of \(1\). Training was performed on a single GPU (NVIDIA A5000), with TF32 acceleration enabled for CUDA operations.

\subsection{Optimization Loss Functions}

To fine-tune the SpeechLLM for rubric-based speech assessment, five modeling strategies of increasing complexity were explored depending on how the task was formulated. Each formulation defines a distinct mapping between the model output and the target scores, influencing both learning behavior and interpretability. In what follows, we describe the loss functions used for the classification and regression variants of our experiments.

\subsubsection{Discrete Classification (DiCl)}

This method served as our baseline for sentence-level assessment. Each speech rubric was formulated as a five-class classification task, where every utterance was assigned one of five categorical levels: Very Poor, Poor, Fair, Good, or Very Good. 
To obtain these labels, we discretize the original 1--10 human rater scores into five ordinal categories using the rubric defined by \citet{zhang21x_interspeech}. Scores of 1–2 were mapped to Very Poor, 3–4 to Poor, 5–6 to Fair, 7–8 to Good, and 9–10 to Very Good. 
A softmax output layer produced the probability distribution over the five classes. The model was optimized using the standard cross-entropy loss:
\[
\mathcal{L}_{\text{DiCl}} = 
\frac{1}{N} \sum_{i=1}^{N}
\left[
- \sum_{c=1}^{C} y_{i,c} \log(\hat{y}_{i,c})
\right]
\]
where \(N\) is the number of utterances in the dataset, \(i\) indexes each utterance, \(C\) denotes the number of classes, \(y_{i,c}\) is the one-hot ground-truth label, and \(\hat{y}_{i,c}\) is the predicted probability for class \(c\).

This loss treats all classes as independent and penalizes all misclassifications equally. 
For instance, predicting Very Good when the true label is Very Poor incurs the same penalty as predicting Good instead of Fair.

\subsubsection{Single Rubric Regression with Mean Squared Error (SRR.M)}

In this setting, the sentence-level assessment was formulated as a regression task, where the model predicts a continuous score within the range [1, 10] for each rubric independently. Compared to the classification approach, regression provides finer granularity, as the MSE loss penalizes predictions in proportion to their numerical deviation from the true score. For example, predicting 7.5 for a true score of 8 incurs a smaller penalty than predicting 5 for 8, thereby preserving ordinal relationships and avoiding discretization artifacts.

A separate regression head was used to predict a single continuous value for each aspect (accuracy, fluency, or prosody). The model was trained to minimize the MSE loss:
\[
\mathcal{L}_{\text{SRR.M}} = 
\frac{1}{N} \sum_{i=1}^{N} (y_i - \hat{y}_i)^2
\]
where \(N\) is the total number of utterances, \(i\) indexes each utterance, \(y_i\) is the gold mean score per aspect (averaged over the five human raters), and \(\hat{y}_i\) is the model-predicted score for the same sample.

\subsubsection{Multi Rubric Regression with Mean Squared Error (MRR.M)}
In this configuration, the assessment task was extended to a multi-output regression problem, where the model simultaneously predicts continuous scores for all three rubrics. 
This was achieved using a shared encoder followed by three parallel regression heads, each producing one scalar output per aspect. It also improves computational efficiency, as a single model produces a structured evaluation vector \([\hat{y}_{\text{acc}}, \hat{y}_{\text{flu}}, \hat{y}_{\text{pro}}]\) without requiring separate fine-tuning for each rubric.
The training objective minimizes the average MSE across all three rubrics:
\[
\mathcal{L}_{\text{MRR.M}} = 
\frac{1}{3} \sum_{\text{aspect}} 
\frac{1}{N} \sum_{i=1}^{N} 
\left(y_{i,\text{aspect}} - \hat{y}_{i,\text{aspect}}\right)^2
\]
where \(N\) is the total number of utterances, \(i\) indexes each utterance, and \(y_{i,\text{aspect}}\) and \(\hat{y}_{i,\text{aspect}}\) denote the gold and predicted mean scores, respectively, for each rubric.

\subsubsection{Multi Rubric Regression with Gaussian Negative Log-likelihood (MRR.G)}

Building upon the multi-head regression framework, this variant introduces uncertainty estimation by allowing each output head to predict not only the mean score \(\mu_i\) but also the corresponding variance \(\sigma_i^2\) for every rubric, accuracy, fluency, and prosody. This formulation enables the model to express both its prediction and its confidence for each utterance. 
Fine-tuning employed the GNLL loss, which penalizes large prediction errors while accounting for the predicted uncertainty. 

For each utterance, the mean of the five rater scores was used as the gold standard, making this a single-target regression task that reflects the central human consensus. 
The GNLL loss for each aspect was defined as:
\[
\mathcal{L}_{\text{MRR.G}}(i)
= \frac{(\bar{y}_i - \mu_i)^2}{2\,\sigma_i^2}
+ \frac{1}{2}\log\sigma_i^2
\]
where \(\bar{y}_i\) denotes the averaged human rating, and \((\mu_i, \sigma_i^2)\) are the predicted mean and variance, respectively. 
The total loss was computed as the mean across the three rubrics:
\[
\mathcal{L}_{\text{total}} = \frac{1}{3} \sum_{\text{aspect}} \frac{1}{N}\sum_{i=1}^{N} \mathcal{L}_{\text{MRR.G}}(i)
\]

This formulation enables the model to capture aleatoric uncertainty by adjusting its predicted variance \(\sigma_i^2\) according to sample difficulty.

\subsubsection{Multi Rubric Multi Rater Regression with Gaussian Negative Log-likelihood and Conformer Prediction (MRR.GC)}

This configuration extends the uncertainty-aware regression framework by directly modeling all five human rater scores per utterance, rather than relying solely on their mean. 
For each rubric, accuracy, fluency, and prosody, the model predicts both a mean score \(\mu_i\) and a variance \(\sigma_i^2\), jointly capturing the central tendency and spread of human judgments. 
Fine-tuning again employed the GNLL loss, which in this case integrates the inter-rater variance term \(s_i^2\) to account for disagreement among raters:
\[
\mathcal{L}_{\text{MRR.GC}} = 
\frac{1}{3N} \sum_{\text{aspect}} \sum_{i=1}^{N}
\left[
\frac{(\bar{y}_i - \mu_i)^2 + s_i^2}{2\,\sigma_i^2}
+ \frac{1}{2}\log\sigma_i^2
\right]
\]
where \(N\) is the total number of utterances, \(\bar{y}_i = \frac{1}{R}\sum_{r=1}^{R} y_{i,r}\) is the mean rater score, \(R=5\) is the number of raters, and
\[
s_i^2 = \frac{1}{R}\sum_{r=1}^{R}(y_{i,r} - \bar{y}_i)^2
\]
represents the inter-rater variance for each utterance \(i\). 
This formulation explicitly incorporates rater disagreement into the loss, enabling the model to reflect both prediction uncertainty and observed human variability.

After fine-tuning, conformal calibration was applied using a 5-fold split of the test set to empirically calibrate the predictive intervals. 
Normalized residuals \(|y_i - \mu_i| / \sigma_i\) were computed on the calibration folds to estimate aspect-wise quantiles (\(q_{\text{aspect}}\)) corresponding to a target coverage of 90\%. 
Final prediction intervals were then obtained as \([\mu_i - q_{\text{aspect}}\sigma_i,\; \mu_i + q_{\text{aspect}}\sigma_i]\).

This combination of multi-rater supervision and conformal prediction allows the model to capture both inter-rater variability and prediction uncertainty in a statistically interpretable manner. Conformal calibration further ensures empirical coverage, guaranteeing that a defined proportion of true scores fall within the predicted confidence intervals.

\subsection{Evaluation Metrics}

The model performance was assessed using a comprehensive set of metrics reflecting both categorical decision quality and numerical agreement with human ratings. For the DiCl baseline setup, performance was assessed using the Weighted F1-score and the Matthews Correlation Coefficient (MCC). Weighted F1 accounts for class imbalance by computing the F1-score per class and averaging it by class frequency, while MCC quantifies the overall correlation between predicted and true labels, providing a balanced measure even under uneven label distributions.

For all four regression-based methods (SRR.M, MRR.M, MRR.G, and MRR.GC), we report five complementary metrics: Weighted F1, MCC, Pearson Correlation Coefficient (PCC), Root MSE (RMSE), and Quadratic Weighted Kappa (QWK). Together, these capture both categorical agreement and continuous correlation with human ratings. The predicted continuous scores were rounded to the nearest integer on the 1--10 scale for computing Weighted F1 and MCC, ensuring comparability with discrete human ratings. PCC measures the linear association between predicted and gold scores, and RMSE quantifies the average prediction error. Quadratic Weighted Kappa (QWK) measures the agreement between two ratings on an ordinal scale, accounting for chance agreement and the distance between rating categories. It ranges from -1 (worse than chance) to 1 (perfect agreement), with 0 indicating random agreement. QWK penalizes larger score discrepancies more heavily than smaller ones, making it well-suited for ordinal rating tasks. For QWK (M–R), the agreement is computed between the model and each of the five human raters individually and reported as mean $\pm$ standard deviation across raters. 

The regression results were analyzed under two assessment conditions: 
(1) a strict, exact-match setting without tolerance, and (2) a relaxed setting that allows a $\pm1$ score tolerance to account for natural rater variability. In practice, small differences, such as one rater giving a 7 and another an 8, are not considered true disagreements but normal variations in human judgment. Prior work in speaking and writing assessment (e.g., TOEFL, SpeechRater) reports inter-rater standard deviations of roughly 0.5–1.0 on a 10-point scale, supporting this tolerance as a realistic estimate of human rating variability. For the MRR.GC configuration, we also calculated the percentage of human scores falling within the model's predicted High–Low confidence range for each aspect. This coverage metric reflects how well the model’s predictive intervals capture real human variability, serving as a direct indicator of calibration quality. Because QWK depends on exact ordinal matches, it is computed only under the strict setting, not with the ±1 tolerance.

\section{Results}

\subsection{Inter-Rater Reliability QWK (R-R)}
Before presenting the results of the five models, we first report the inter-rater reliability (QWK, R–R) in Table~\ref{tab:rr-qwk}. The shown mean QWK values are averaged across all ten possible rater pairs, indicating overall moderate agreement among raters. 

\begin{table}[!htbp]
\centering
\resizebox{\columnwidth}{!}{%
\begin{tabular}{lccc}
\toprule
 & \textbf{Accuracy} & \textbf{Fluency} & \textbf{Prosody} \\
\midrule
QWK (R--R) & 0.5585~$\pm$~0.0671 & 0.5019~$\pm$~0.1353 & 0.5021~$\pm$~0.1153 \\
\bottomrule
\end{tabular}}
\caption{QWK (mean~$\pm$~SD) across human raters.}
\label{tab:rr-qwk}
\end{table}

\subsection{Classification-Based Assessment}

Table~\ref{tab:classification-results2} summarizes the baseline classification performance (DiCl) across the three assessment rubrics: accuracy, fluency, and prosody. 
Overall, DiCl results demonstrate moderate discriminative ability across rubrics, providing a reference point for subsequent regression-based approaches that aim to capture finer score variations. 


\begin{table}[!htbp]
\centering
\footnotesize
\setlength{\tabcolsep}{10pt}
\begin{tabular}{lcc}
\toprule
\textbf{Rubrics} & \textbf{$\uparrow$ F1} & \textbf{$\uparrow$ MCC} \\
\midrule
Accuracy & 0.558 & 0.341 \\
Fluency  & 0.642 & 0.452 \\
Prosody  & 0.670 & 0.468 \\
\bottomrule
\end{tabular}
\caption{Classification performance across rubrics for DiCl. Arrows indicate that higher is better.}
\label{tab:classification-results2}
\end{table}

\begin{figure}[!htbp]
  \centering
  \includegraphics[width=\linewidth,trim=05 05 10 10,clip]{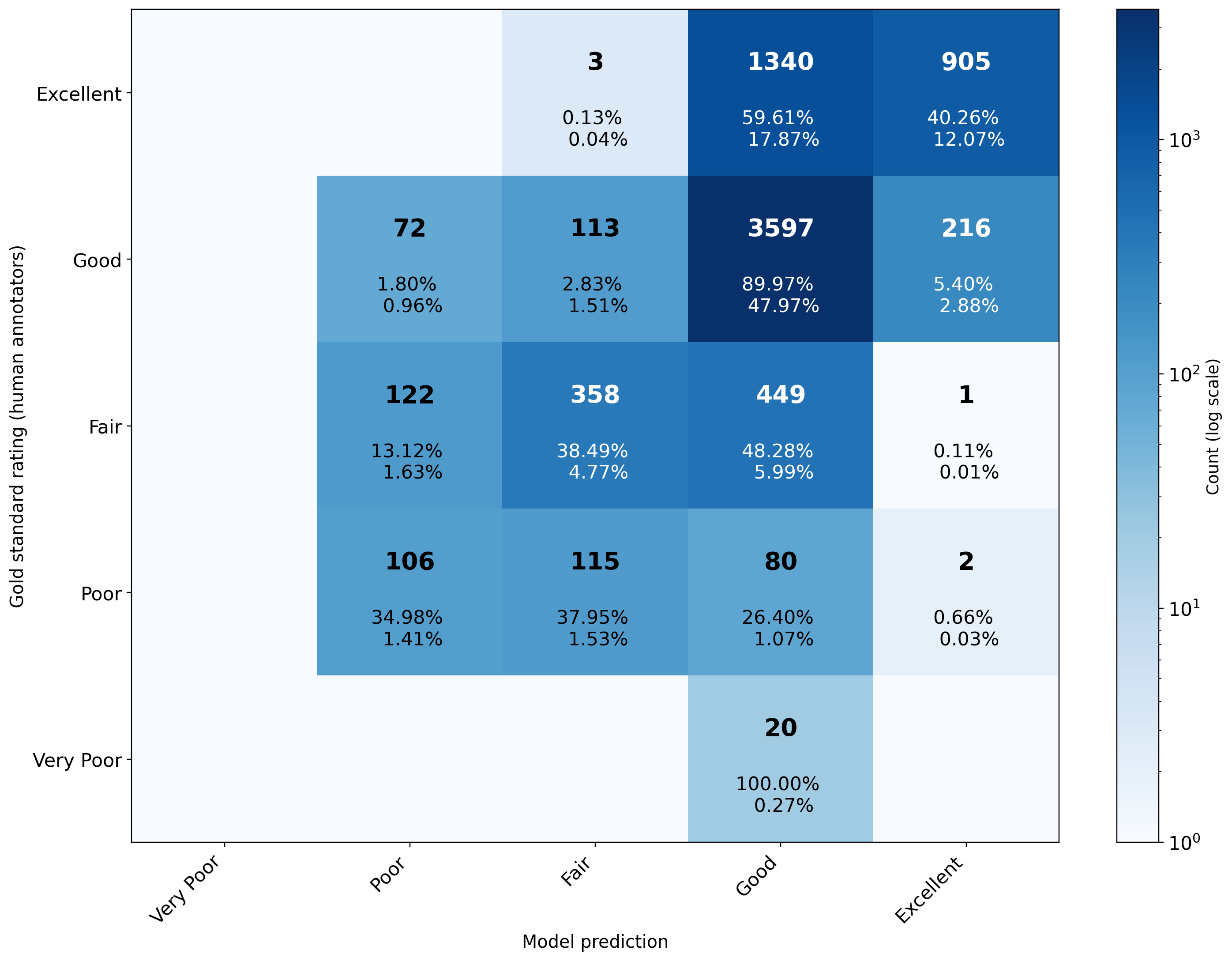}
  \caption{Aggregated confusion matrix across all rubrics for DiCl. Rows show human (gold-standard) ratings. Columns show model predictions. Each cell displays the count (top), percentage within the true class (middle), and percentage across all samples (bottom). }
  \label{fig:conf-matrix-class}
\end{figure}

Figure~\ref{fig:conf-matrix-class} shows the aggregated confusion matrix across all rubrics for DiCl. Predictions are largely concentrated along the main diagonal, indicating good alignment between the model's output and human ratings. The Good category dominates the predictions.
Misclassification mainly occurs between adjacent levels (see Figure~\ref{fig:conf-matrix-class}).


\subsection{Regression-Based Assessment}

In this section, we compare four regression-based fine-tuning configurations: SRR.M, MRR.M, MRR.G, and MRR.GC. Performance is reported for each rubric (accuracy, fluency, and prosody) under both strict (exact-match) and lenient evaluation settings ($\pm1$ tolerance or High–Low calibration).

\subsubsection{Exact-Match Regression Results (without tolerance)}

Table~\ref{tab:regression-results} presents the regression-based results under exact-match evaluation. 
Performance is reported separately for the three rubrics: accuracy, fluency, and prosody, to analyze aspect-specific trends before summarizing overall behavior across model variants. 

\begin{table}[!htbp]
\centering
\setlength{\tabcolsep}{2.5pt}
\renewcommand{\arraystretch}{0.82}
\setlength{\cmidrulewidth}{0.25pt}

\scriptsize
\begin{tabular}{l l c c c c l}
\toprule
 & \textbf{Rubrics}  & \textbf{$\uparrow$ F1} & \textbf{$\uparrow$ MCC} & \textbf{$\uparrow$ PCC} & \textbf{$\downarrow$ RMSE} & \textbf{$\uparrow$ QWK (M--R)} \\
\midrule

\multirow{6}{*}{\rotlabel{\textbf{SRR.M}}}
 & Accuracy              & \num{0.2479} & \num{0.0843} & \num{0.7390} & \num{1.1854} & \multirow{2}{*}{\num{0.432}~$\pm$~\num{0.046}} \\
 & $\pm$1                & \num{0.7966} & \num{0.7543} & \num{0.9417} & \num{0.5456} & \\
\aspectsep
 & Fluency               & \num{0.3717} & \num{0.2416} & \num{0.7956} & \num{0.9326} & \multirow{2}{*}{\num{0.445}~$\pm$~\num{0.081}} \\
 & $\pm$1                & \num{0.9137} & \num{0.9000} & \num{0.9571} & \num{0.4218} & \\
\aspectsep
 & Prosody               & \num{0.4073} & \num{0.2838} & \num{0.7728} & \num{0.9163} & \multirow{2}{*}{\num{0.468}~$\pm$~\num{0.093}} \\
 & $\pm$1                & \num{0.9216} & \num{0.9117} & \num{0.9558} & \num{0.4357} & \\
\midrule

\multirow{6}{*}{\rotlabel{\textbf{MRR.M}}}
 & Accuracy              & \num{0.2424} & \num{0.0710} & \num{0.6969} & \num{1.2894} & \multirow{2}{*}{\num{0.497}~$\pm$~\num{0.051}} \\
 & $\pm$1                & \num{0.7561} & \num{0.6830} & \num{0.7454} & \num{0.6221} & \\
\aspectsep
 & Fluency               & \num{0.4571} & \num{0.3161} & \num{0.7522} & \num{0.8910} & \multirow{2}{*}{\num{0.501}~$\pm$~\num{0.078}} \\
 & $\pm$1                & \num{0.9378} & \num{0.9286} & \num{0.8890} & \num{0.4462} & \\
\aspectsep
 & Prosody               & \num{0.3435} & \num{0.1918} & \num{0.7463} & \num{1.0342} & \multirow{2}{*}{\num{0.527}~$\pm$~\num{0.052}} \\
 & $\pm$1                & \num{0.8809} & \num{0.8536} & \num{0.8457} & \num{0.4553} & \\
\midrule

\multirow{6}{*}{\rotlabel{\textbf{MRR.G}}}
 & Accuracy              & \num{0.2285} & \num{0.0705} & \num{0.7352} & \num{1.1762} & \multirow{2}{*}{\num{0.464}~$\pm$~\num{0.049}} \\
 & $\pm$1                & \num{0.7785} & \num{0.7428} & \num{0.8444} & \num{0.5402} & \\
\aspectsep
 & Fluency               & \num{0.3931} & \num{0.2657} & \num{0.8003} & \num{0.9237} & \multirow{2}{*}{\num{0.463}~$\pm$~\num{0.054}} \\
 & $\pm$1                & \num{0.9179} & \num{0.9078} & \num{0.8489} & \num{0.4308} & \\
\aspectsep
 & Prosody               & \num{0.4095} & \num{0.2821} & \num{0.7933} & \num{0.9082} & \multirow{2}{*}{\num{0.494}~$\pm$~\num{0.079}} \\
 & $\pm$1                & \num{0.9226} & \num{0.9133} & \num{0.8526} & \num{0.4185} & \\
\midrule

\multirow{6}{*}{\rotlabel{\textbf{MRR.GC}}}
 & Accuracy              & \num{0.4014} & \num{0.2654} & \num{0.7649} & \num{1.0159} & \multirow{2}{*}{\num{0.505}~$\pm$~\num{0.039}} \\
 & High--Low Cal         & \num{0.8707} & \num{0.8539} & \num{0.8716} & \num{0.5467} & \\
\aspectsep
 & Fluency               & \num{0.4589} & \num{0.3397} & \num{0.8521} & \num{0.7768} & \multirow{2}{*}{\num{0.521}~$\pm$~\num{0.022}} \\
 & High--Low Cal         & \num{0.9276} & \num{0.9176} & \num{0.9164} & \num{0.4212} & \\
\aspectsep
 & Prosody               & \num{0.4767} & \num{0.3458} & \num{0.8361} & \num{0.7903} & \multirow{2}{*}{\num{0.496}~$\pm$~\num{0.085}} \\
 & High--Low Cal         & \num{0.9319} & \num{0.9222} & \num{0.9119} & \num{0.3952} & \\
\bottomrule
\end{tabular}%
\caption{Regression-based results across all model families. 
Arrows indicate the direction of improvement. 
Tolerance (\(\pm1\)) and High--Low Calibration indicate lenient evaluation settings where predictions within the tolerance or predicted uncertainty range are considered acceptable.}
\label{tab:regression-results}
\end{table}

Across all three rubrics, Table~\ref{tab:regression-results} results indicate consistent improvements with increasing model complexity (i.e. from top to bottom). 
The multi-head models (MRR.M, MRR.G, MRR.GC) generally outperform single-head regression (SRR.M), showing the benefit of shared representation learning across rubrics. 
Introducing the GNLL objective (MRR.G) further improves robustness by jointly modeling prediction uncertainty, and the full MRR.GC configuration achieves the strongest overall performance and highest alignment with human ratings across all evaluation metrics.

\begin{figure*}[t]
  \centering
  \includegraphics[width=\textwidth]{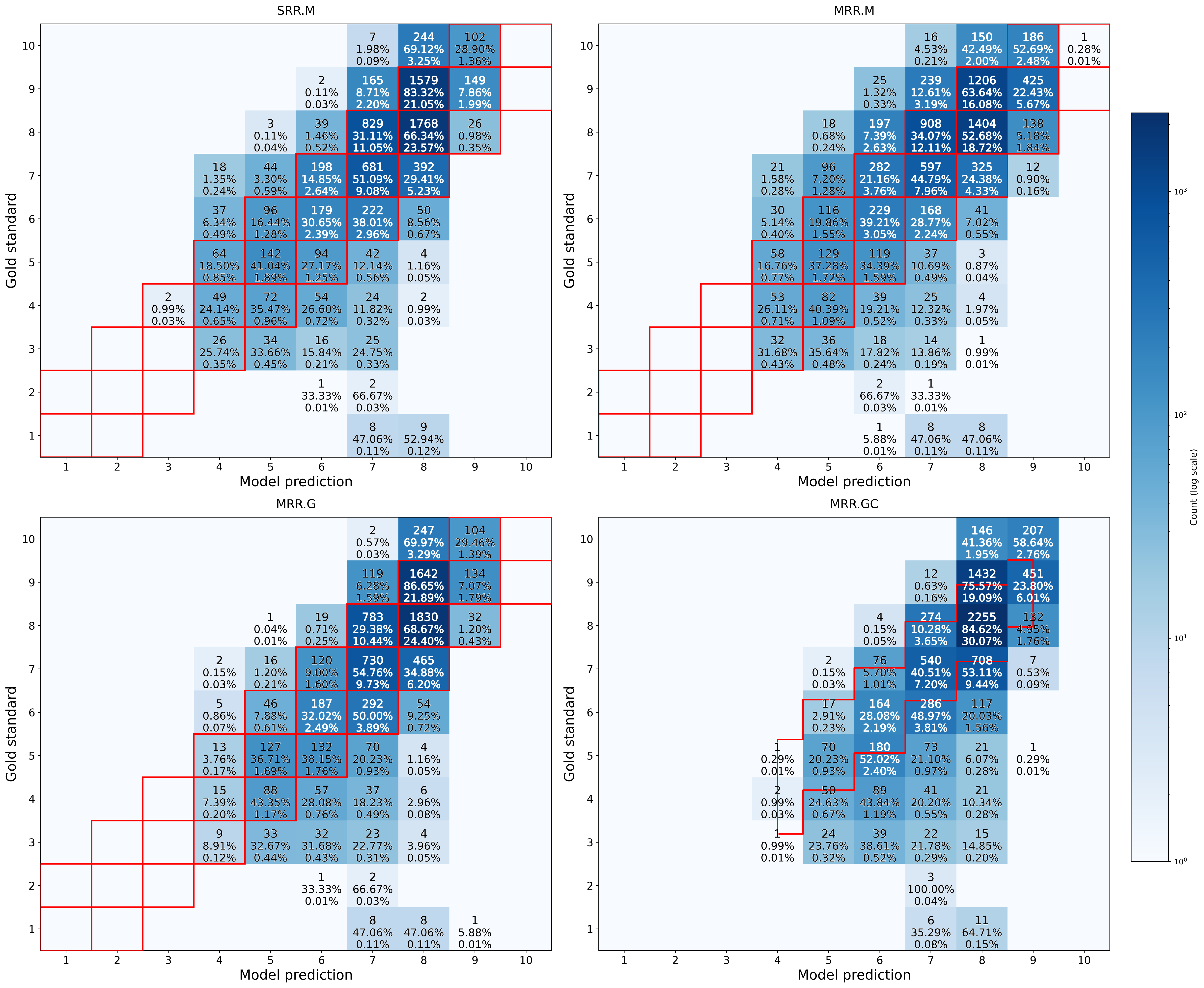}
  \caption{Aggregated confusion matrices across regression methods (SRR.M, MRR.M, MRR.G, MRR.GC). Red boxes in the first three panels indicate the $\pm1$ tolerance region, while in the bottom-right panel, red contours denote the median calibrated range from conformal prediction. Each cell shows the total count (top), the percentage within the true class (middle), and the percentage relative to all samples (bottom).}
  \label{fig:wide}
\end{figure*}

Figure~\ref{fig:wide} presents the confusion matrices for all regression configurations.
Overall, all models show tight clustering around mid-level utterances but limited separability at the lowest and highest scores.
Predictions across models are concentrated in the mid-range (scores~5--8), while the extremes (1--2 and~10) are rarely or never predicted.

\subsubsection{Regression Results with Tolerance and Calibration}

Table~\ref{tab:regression-results} also summarizes the regression-based results evaluated under a relaxed setting that allows a tolerance of $\pm 1$~score point and, in the final configuration, incorporates conformal calibration. 
Under this evaluation, predictions within one score point of the human rating are considered 
acceptable, reflecting natural variability among human raters. 
When lenient settings are applied, the overall 
performance increases across all rubrics. 
The $\pm 1$~tolerance evaluation substantially boosts F1, MCC, and PCC scores, reflecting stable prediction behavior within one rating level of the human mean. 
In the uncertainty-aware MRR.GC model, conformal calibration achieves a comparable improvement by explicitly modeling confidence intervals instead of tolerance bands. 

Figure~\ref{fig:wide} also illustrates the confusion matrices under lenient evaluation. 
In the first three panels (SRR.M, MRR.M, and MRR.G), the red boxes mark the $\pm 1$~tolerance region 
around the diagonal, highlighting predictions within one score point of the human gold standard. 
The bottom-right panel (MRR.GC) shows calibrated boundaries from conformal prediction, with red contours marking the median empirical high–low range per score bin.




\subsubsection{Coverage Analysis under Conformal Calibration} 
To evaluate how well the calibrated prediction intervals align with the empirical variability among human ratings, we present in Table~\ref{tab:raters-in-range} the percentage of utterances for which a given number of human raters (\(0\)–\(5\)) fall within the model's predicted high–low interval, obtained from conformal calibration in the \textsc{MRR.GC} configuration.

\begin{table}[!htbp]
\centering
\resizebox{\columnwidth}{!}{%
\begin{tabular}{cccc}
\toprule
\textbf{$\leq N$ raters} & \textbf{Accuracy (\%)} & \textbf{Fluency (\%)} & \textbf{Prosody (\%)} \\
\midrule
5 & 6.68 & 0.96 & 1.92 \\
4 & 25.92 & 8.60 & 12.72 \\
3 & 60.04 & 34.24 & 44.16 \\
2 & 83.92 & 67.20 & 77.56 \\
1 & 93.84 & 91.52 & 94.68 \\
\bottomrule
\end{tabular}}
\caption{Cumulative percentage with at most $N$ raters within the model's prediction interval (monotonic increasing with $N$).}
\label{tab:raters-in-range}
\end{table}

\section{Discussion}



In this section, we first discuss the main findings and then provide an answer to our RQ. To address the RQ, we fine-tuned the Qwen2-Audio-7B-Instruct SpeechLLM using rubric-guided data for multi-aspect assessment of L2 read speech under five different configurations. We observe that performance gradually improves from SRR.M to MRR.M, MRR.G, and MRR.GC (Table \ref{tab:regression-results}). Multi-rubric regression improves stability and captures cross-rubric dependencies. Incorporating the GNLL objective in MRR.G further improves calibration by modeling variability in human ratings, in line with computer vision research \citep{kendall2017uncertainties}. The final MRR.GC, achieves the best overall alignment with human judgments by combining multi-rater supervision and conformal calibration, as proven in other research fields \citep{braun2025multivariate}, which produces adaptive confidence intervals that reflect empirical rater disagreement.

Allowing a tolerance of $\pm1$~score point improves F1, MCC, and PCC, indicating that minor discrepancies fall within normal perceptual variability (Table~\ref{tab:regression-results}, red boxes in Fig.~\ref{fig:wide}). However, this margin spans 20\% of the 10-point scale, so part of the gain stems from more lenient evaluation. In contrast, conformal calibration in MRR.GC provides a principled way for quantifying uncertainty: interval widths adapt to local variability, yielding statistically valid confidence bounds consistent with human rating behavior. As shown in Table~\ref{tab:raters-in-range}, higher coverage for accuracy suggests closer human–model agreement, while narrower intervals for fluency and prosody indicate more consistent rater behavior and tighter calibration around consensus scores.

Human raters show moderate agreement among themselves (Table~\ref{tab:rr-qwk}). Furthermore, we observed higher inter-rater reliability for accuracy compared to fluency and prosody (Table~\ref{tab:rr-qwk}) and the percentages of human ratings that fall within the model’s predicted high–low intervals (Table~\ref{tab:raters-in-range}). On the other hand, Table~\ref{tab:regression-results} shows that the regression-based results are better for fluency and prosody than for accuracy. Possible explanations for these differences might lie in (a) the definitions of the three aspects of accuracy, fluency, and prosody and (b) the operationalization of these three constructs. As to (a), while the definition of accuracy seems rather straightforward, the correctness of pronunciation, the definitions of fluency and prosody are somewhat confusing. As a matter of fact, they seem to mix several features. For example, speaking rate could just as well be part of fluency instead of prosody, based on the definition of fluency as temporal smoothness. As to (b), the instructions for fluency and prosody may be easier to operationalize for LLMs than those for accuracy. Temporal aspects have long been recognized as easier to compute automatically than features related to segmental quality that may involve multiple dimensions. Previous research on automatic assessment of non-native speech revealed that ASR systems are better at capturing temporal-related aspects of non-native speech than those related to segmental quality \citep{cucchiarini2000quantitative,cucchiarini2000different,cucchiarini2002quantitative}.


The main limitations of our work concern data imbalance and generalization. Model predictions tend to cluster around scores 6 to 8, reflecting the skewed distribution of human ratings and reducing sensitivity to extreme proficiency levels. This mid-range bias inflates global performance metrics and constrains the model's ability to generalize to underrepresented proficiency extremes.

Finally, in response to our RQ, the results presented in the current paper demonstrate that the proposed models align closely with human judgments across all rubrics, indicating that a well-designed SpeechLLM can effectively support multi-aspect automatic assessment. Among them, the MRR.GC model performs best, offering the additional advantage of capturing not only mean human ratings but also their variability.




\section{Conclusion}

This study investigated whether a rubric-guided SpeechLLM can approximate human judgments in the assessment of L2 reading speech. To systematically examine whether incorporating additional configurations can improve robustness and alignment with human judgments, we developed five SOTA modeling strategies of increasing complexity for fine-tuning the Qwen2-Audio-7B-Instruct SpeechLLM,  based on insights from related literature. Among the evaluated strategies, MRR.GC achieved the strongest overall alignment with human raters, with aggregated performance across Accuracy, Fluency, and Prosody of $\text{PCC}\approx0.81$, $\text{RMSE}\approx0.83$, and $\text{QWK}\approx0.50$. These results demonstrate that incorporating multi-rater supervision, Gaussian uncertainty modeling, and conformal calibration yields reliable and interpretable scoring for multi-aspect L2 reading speech assessment. However, the model behaves conservatively at score extremes, highlighting the need to address mid-range bias and extend the framework to diagnostic feedback and error localization to improve assessment validity. Future work will extend beyond scoring to diagnostic feedback and error localization for actionable learner guidance.

\section{Acknowledgements}
This publication is part of the project Responsible AI for Voice Diagnostics (RAIVD) with file number NGF.1607.22.013 of the research programme NGF AiNed Fellowship Grants which is financed by the Dutch Research Council (NWO).

\section{Bibliographical References}\label{sec:reference}

\bibliographystyle{lrec2026-natbib}
\bibliography{lrec2026-example}

\end{document}